\documentclass[conference]{IEEEtran}
\IEEEoverridecommandlockouts
\usepackage{cite}
\usepackage{amsmath,amssymb,amsfonts}
\usepackage{algorithmic}
\usepackage{graphicx}
\usepackage{textcomp}
\usepackage{xcolor}
\def\BibTeX{{\rm B\kern-.05em{\sc i\kern-.025em b}\kern-.08em
    T\kern-.1667em\lower.7ex\hbox{E}\kern-.125emX}}

\usepackage[utf8]{inputenc}
\usepackage{cleveref}
\usepackage{booktabs}
\usepackage{adjustbox}
\usepackage{threeparttable}
\usepackage{pifont}
\usepackage{url}

\newcommand{\e}[1]{{\mathbb E}\left[ #1 \right]}
\usepackage[ruled,vlined, linesnumbered, commentsnumbered]{algorithm2e}
\usepackage{graphicx}
\usepackage{float}
\usepackage{caption}
\usepackage{subcaption}
\usepackage{multirow}
\usepackage{tikz}

\DeclareMathAlphabet{\mathcal}{OMS}{cmsy}{m}{n}

\usepackage[nolist,nohyperlinks]{acronym}
\newacro{CNN}{Convolutional Neural Network}
\newacro{CNNs}{Convolutional Neural Networks}
\newacro{DNN}{Deep Neural Network}
\newacro{GPS}{Global Positioning System}
\newacro{GNSS}{Global Navigation Satellite System}
\newacro{NLOS}{non-line-of-sight}
\newacro{ADAS}{Advanced Driver Assistance Systems}
\newacro{LIDAR}[LiDAR]{Light Detection And Ranging}
\newacro{HD map}{High Definition map}
\newacro{EV}{Embedding Vector}
\newacro{SLAM}{Simultaneouos Localization And Mapping}
\newacro{MLP}{MultiLayer Perceptron}
\newacro{IMU}{Inertial Measurement Unit}
\newacro{ML}{Machine Learning}
\newacro{SfM}{Structure from Motion}
\newacro{PnP}{Perspective-n-Points}
\newacro{ASPP}{Atrous Spatial Pyramid Pooling}

\newacro{NN}{Neural Network}
\newacro{NNs}{Neural Networks}
\newacro{MCD}{Monte Carlo Dropout}
\newacro{DE}{Deep Ensemble}
\newacro{DEs}{Deep Ensembles}
\newacro{DER}{Deep Evidential Regression}
\newacro{NIG}{Normal Inverse Gamma}
\newacro{NLL}{Negative Log Likelihood}
\newacro{ER}{Evidence Regularizer}
\newacro{FC}{Fully Connected}
\newacro{EKF}{Extended Kalman Filter}
\newacro{ATE}{Absolute Trajectory Error}
\newacro{RANSAC}{RANdom SAmple Consensus}

\def\etal{\emph{et al. }}
\def\eg{\emph{e.g., }}
\def\ie{\emph{i.e., }}

\def\checkmark{\tikz\fill[scale=0.3](0,.35) -- (.25,0) -- (1,.7) -- (.25,.15) -- cycle;}

\begin{document}

\title{Uncertainty-Aware DNN for Multi-Modal Camera Localization}

\author{\IEEEauthorblockN{Matteo Vaghi$^1$}
\and
\IEEEauthorblockN{Augusto Luis Ballardini$^2$}
\and
\IEEEauthorblockN{Simone Fontana$^1$}
\and
\IEEEauthorblockN{Domenico Giorgio Sorrenti$^1$}
\thanks{$^{1}$Università degli Studi di Milano - Bicocca, Milan, Italy. Corresponding Author: \tt\small m.vaghi9@campus.unimib.it}
\thanks{$^{2}$ Universidad de Alcalá, Alcalá de Henares, Spain.}
}
\IEEEaftertitletext{\vspace{-1.75\baselineskip}}
\maketitle

\begin{abstract}
Camera localization, \ie camera pose regression, represents an important task in computer vision since it has many practical applications such as in the context of intelligent vehicles and their localization. Having reliable estimates of the regression uncertainties is also important, as it would allow us to catch dangerous localization failures. In the literature, uncertainty estimation in \acp{DNN} is often performed through sampling methods, such as \ac{MCD} and \ac{DE}, at the expense of undesirable execution time or an increase in hardware resources. In this work, we considered an uncertainty estimation approach named \ac{DER} that avoids any sampling technique, providing direct uncertainty estimates. Our goal is to provide a systematic approach to intercept localization failures of camera localization systems based on \acp{DNN} architectures, by analyzing the generated uncertainties. We propose to exploit CMRNet, a \ac{DNN} approach for multi-modal image to LiDAR map registration, by modifying its internal configuration to allow for extensive experimental activity on the KITTI dataset. The experimental section highlights CMRNet's major flaws and proves that our proposal does not compromise the original localization performances but also provides, at the same time, the necessary introspection measures that would allow end-users to act accordingly.
\end{abstract}

\begin{IEEEkeywords}
Camera Localization, Deep Learning, Uncertainty estimation
\end{IEEEkeywords}

\section{INTRODUCTION}\label{sec:introduction}

Although DNN-based techniques achieve outstanding results in camera localization \cite{valada_2018, Sarlin_2021_CVPR}, a main challenge is still unsolved: to determine when such models are providing a reliable localization output since inaccurate estimates could endanger other road users. Therefore, being able to assign a reliable degree of uncertainty to the model predictions allows us to decide whether the outputs can be safely used for navigation \cite{McAllister_2017}.
The uncertainty associated with the model output can be of two different types: aleatoric and epistemic.
''Aleatoric uncertainty represents the effect on the output given by variability of the input data that cannot be modeled: this uncertainty cannot be reduced even if more data were to be collected. Epistemic uncertainty, on the other hand, quantifies the lack of knowledge of a model, which arises from the limited amount of data used for tuning its parameters. This uncertainty can be mitigated with the usage of more data.'' Adapted from \cite{kendall_2017_unc}.
DNN-based camera localization proposals that also estimate uncertainty already exist in the literature, \eg \cite{kendall_2016, deng2022deep}. However, only partial comparisons with the consolidated approaches are available, \eg \cite{kendall_2016} just deals with \ac{MCD}. In addition, since those techniques deal only with image data, their effectiveness with multi-modal approaches should be explored.
Given the importance of uncertainty estimation for \ac{DNN}-based camera localization, in this work we propose an application of three state-of-the-art methods for epistemic uncertainty estimation in \ac{CNNs} within a multi-modal camera localization approach, and show that they can provide calibrated uncertainties and that some of them can also be used to detect localization failures.
We chose CMRNet \cite{Cattaneo_2019}, an approach for camera localization using a camera image and an available 3D map, typically built from LiDAR data. The reason is our familiarity with the model and its implementation. Moreover, we consider it significant to have developed a version of a camera localization DNN model that is able to estimate uncertainty by using \ac{DER}.

\begin{figure}
         \centering
         \includegraphics[width=.45\textwidth]{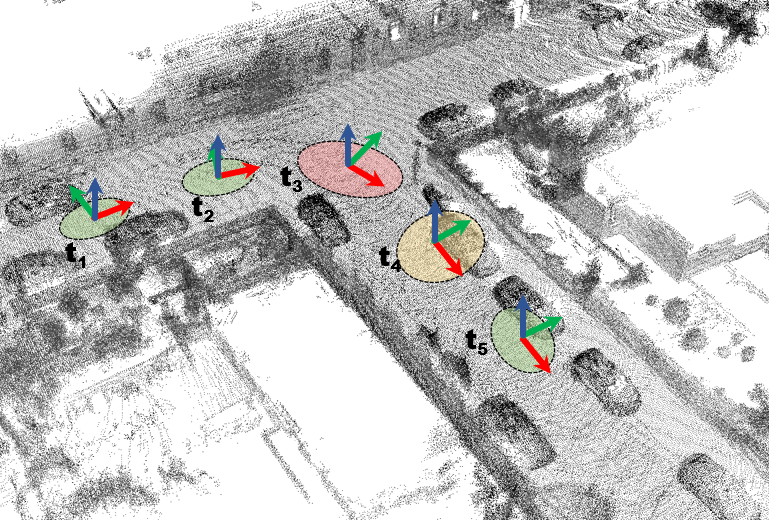}
         \caption{\footnotesize{We compare three approaches for estimating uncertainty in \ac{DNN}s for camera localization by integrating them in a camera-to-LiDAR map registration model. We assess uncertainty quality by measuring calibration, showing that two of the proposed approaches can detect localization failures.}}
         \label{fig:teaser}
         \vspace{-5mm}
\end{figure}


\begin{figure*}
         \centering
         \includegraphics[width=.95\textwidth]{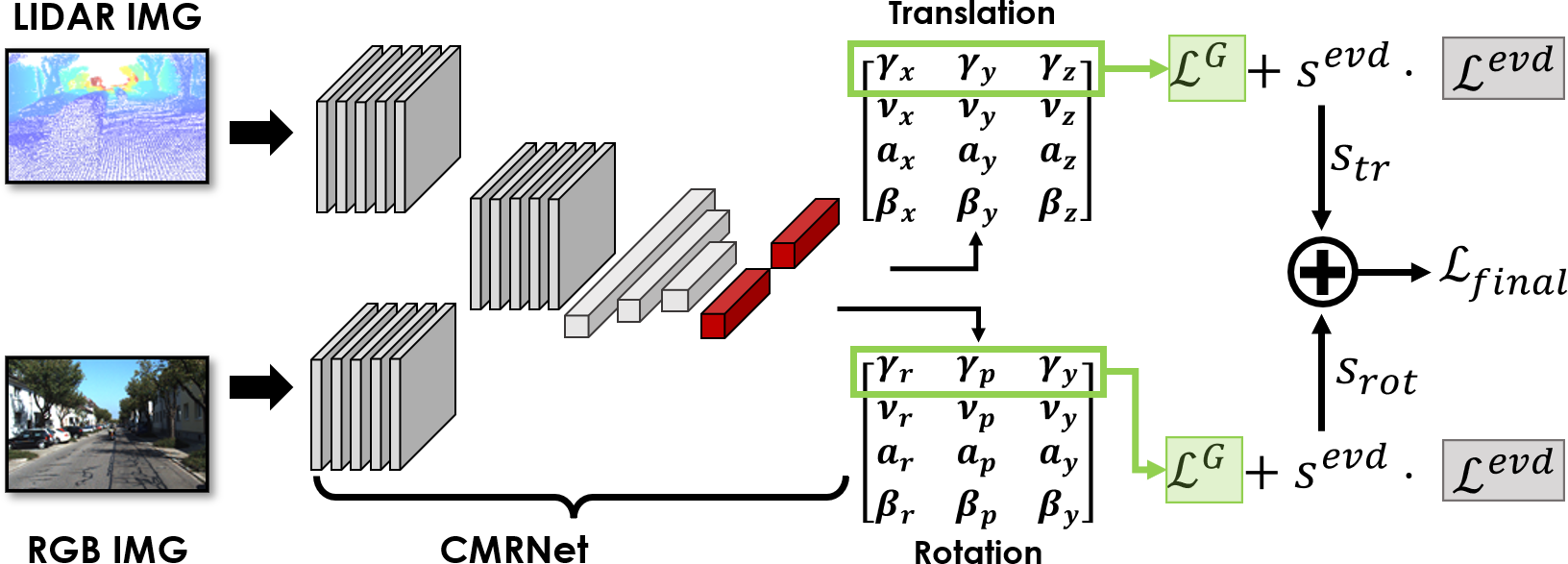}
         \caption{\footnotesize{In this picture the CMRNet + DER approach is shown. The last FC-layers (red) are modified according to the method proposed by Amini \etal \cite{Amini_2020} for estimating the parameters $m_i = (\gamma_i, \nu_i, \alpha_i, \beta_i)$ of different \ac{NIG} distributions. During training, $\mathcal{L}^G$ (green) and $\mathcal{L}^{evd}$ (grey) loss functions are computed both for translation and rotation components.}}
         \label{fig:cmrnet_der}
         \vspace{-0.6cm}
\end{figure*}

\section{RELATED WORK}\label{sec:related}

In the last decade, many \ac{DNN}-based approaches for camera localization emerged.
In general, we can divide existing methods into two categories: camera pose regression \cite{Kendall_2015_ICCV, Kendall_2017_CVPR, valada_2018, Yin_2018_CVPR, Sarlin_2021_CVPR} and place recognition \cite{Arandjelovic_2016_CVPR, zhu_2018, Hausler_2021_CVPR} techniques.
Using an image, the former category predicts the pose of a camera, while the latter finds a correspondence with a previously visited location, depicted in another image.
Multi-modal approaches, which employ image and \ac{LIDAR} data, propose to jointly exploit visual information and the 3D geometry of a scene to achieve higher localization accuracy \cite{Wolcott_2014, Caselitz_2016, Neubert_2017}.
Recently, \ac{DNN}-based methods emerged also for image-to-\ac{LIDAR}-map registration. An example is CMRNet \cite{Cattaneo_2019}, which performs direct regression of the camera pose by implicitly matching RGB images with the corresponding synthetic \ac{LIDAR} image generated using a \ac{LIDAR} map and a rough camera pose estimate. Its ultimate goal is to refine common GPS localization measures. CMRNet is map-agnostic.
Feng \etal \cite{Feng_2019} proposed another multi-modal approach, where a \ac{DNN} is trained to extract descriptors from 2D and 3D patches by defining a shared feature space between heterogeneous data. Localization is then performed by exploiting points for which 2D-3D correspondences have been found. Similarly, Cattaneo \etal \cite{Cattaneo_2020} proposed a \ac{DNN}-based method for learning a common feature space between images and \ac{LIDAR} maps to produce global descriptors, used for place recognition.
Although the previous multi-modal pose regression techniques achieve outstanding results, none of them estimate the epistemic uncertainty of their predictions. This is a severe limitation, especially considering the final goal: to deploy them in critical scenarios, where it is important to detect when the model is likely to fail.
Epistemic uncertainty estimation in \ac{NNs} is a known problem. In the last years, different methods have been proposed to sample from the model posterior \cite{kingma_2015, Balaji_2017} and, more recently, to provide a direct uncertainty estimate through evidential deep learning \cite{Sensoy_2018, Amini_2020, Meinert_2021}.
\ac{NNs} uncertainty estimation gained popularity also in the computer vision field \cite{kendall_2017_unc, Kendall_2018_CVPR}, and different uncertainty-aware camera-based localization approaches have been proposed. For instance, Kendall \etal \cite{kendall_2016} introduced Bayesian PoseNet, a \ac{DNN} that estimates the camera pose parameters and uncertainty by approximating the model posterior employing  dropout sampling \cite{Gal_2016}.
Deng \etal \cite{deng2022deep} proposed another uncertainty-aware model, which relies on Bingham mixture models for estimating a 6DoF pose from an image. Recently, Petek \etal \cite{Petek_2022} proposed an approach to camera localization that exploits an object detection module, which is used to enable localization within sparse HD maps. In particular, their method estimates the vehicle pose using the uncertainty of the objects in the HD map using a \ac{DER} approach \cite{Amini_2020}.
Another interesting approach is HydraNet \cite{peretoukhin_2019}, which is a neural network for estimating uncertainty on quaternions. 
All the mentioned techniques deal with the problem of camera localization using only images, they learn to localize a camera in the environment represented in the training set. In contrast, CMRNet is map-agnostic, \ie by being able to take in input a \ac{LIDAR}-map, it can perform localization also in previously unseen environments.
Furthermore, to the best of our knowledge, this is the first work to implement a \ac{DER}-based approach for direct camera localization.

\section{METHOD}\label{sec:method}
In our analysis of the literature, we could single out three more significant methods for estimating epistemic uncertainty in a \ac{DNN}: \ac{MCD} \cite{Gal_2016}, \ac{DE} \cite{Balaji_2017}, and \ac{DER} \cite{Amini_2020}. Although they all assume that epistemic uncertainty can be described by a normal distribution, they are different techniques and require different interventions on the network to which they are applied.
Therefore, in this section, we first introduce it and then describe the modifications required in CMRNet to estimate uncertainty using each of the three different methods.

\begin{figure*}
     \centering
     \begin{subfigure}[b]{.32\textwidth}
         \centering
         \includegraphics[width=\textwidth]{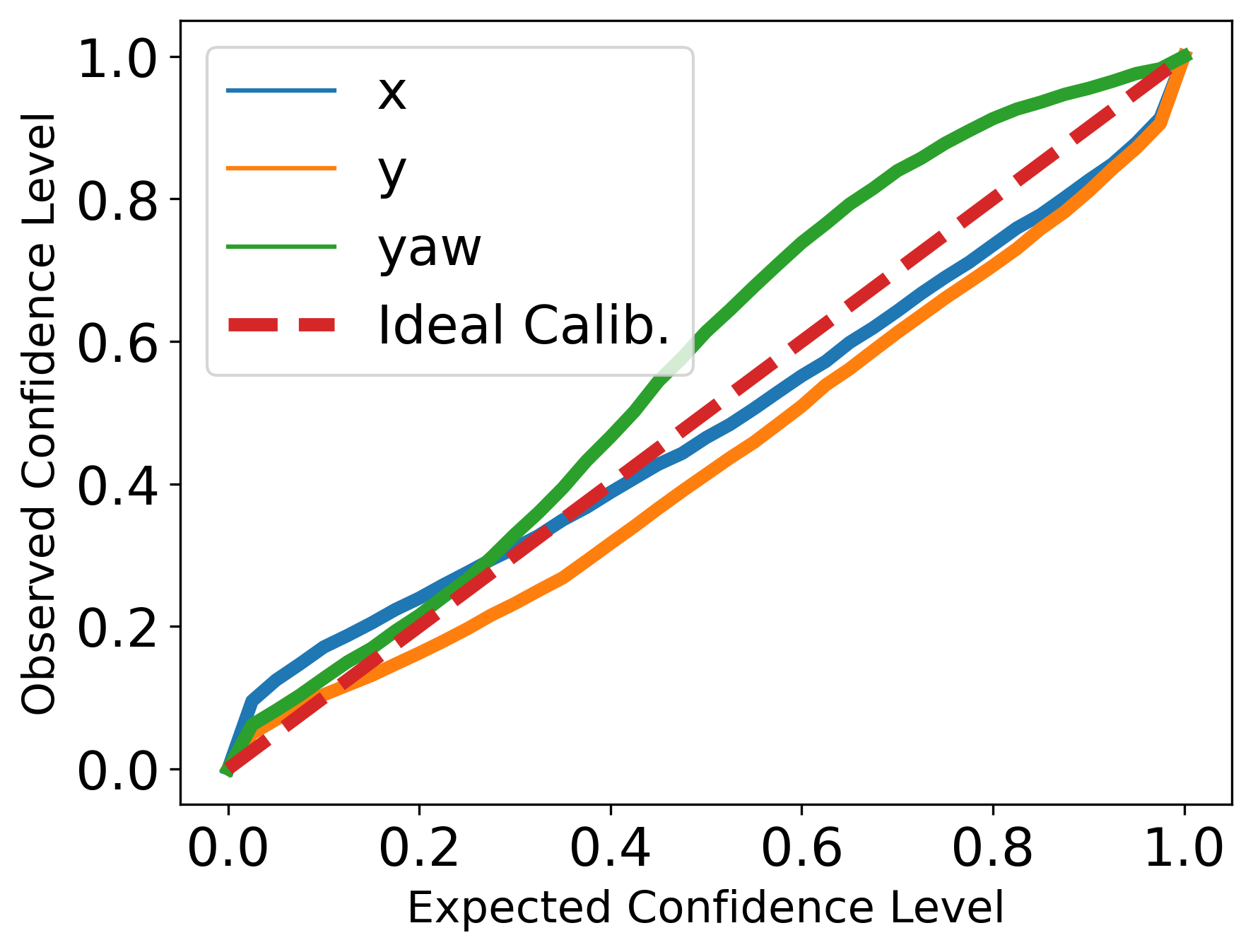}
         \caption{Monte Carlo Dropout}
         \label{fig:cal_mcd}
     \end{subfigure}
     \hfill
     \begin{subfigure}[b]{.32\textwidth}
         \centering
         \includegraphics[width=\textwidth]{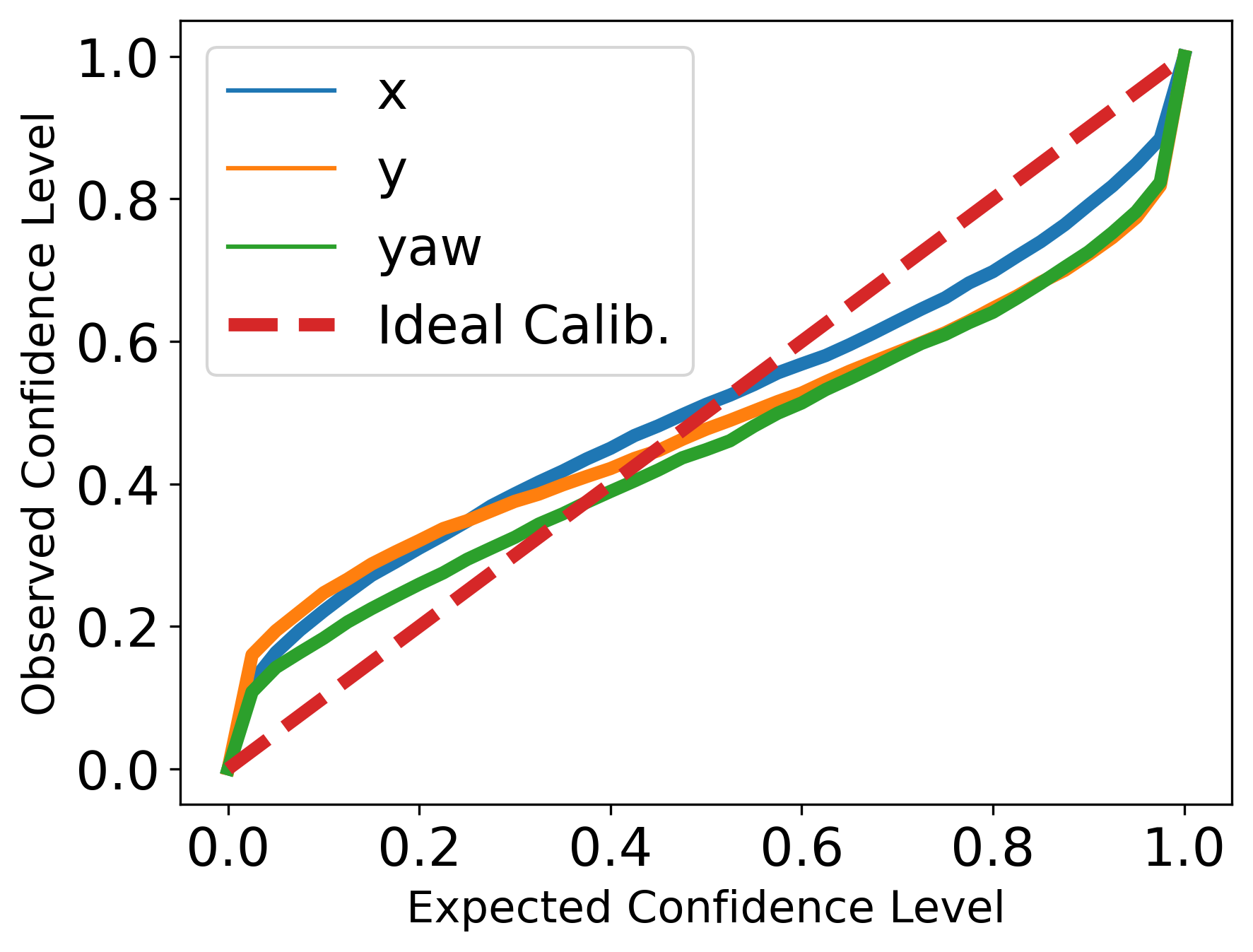}
         \caption{Deep Ensemble}
         \label{fig:cal_de}
     \end{subfigure}
     \hfill
     \begin{subfigure}[b]{.32\textwidth}
         \centering
         \includegraphics[width=\textwidth]{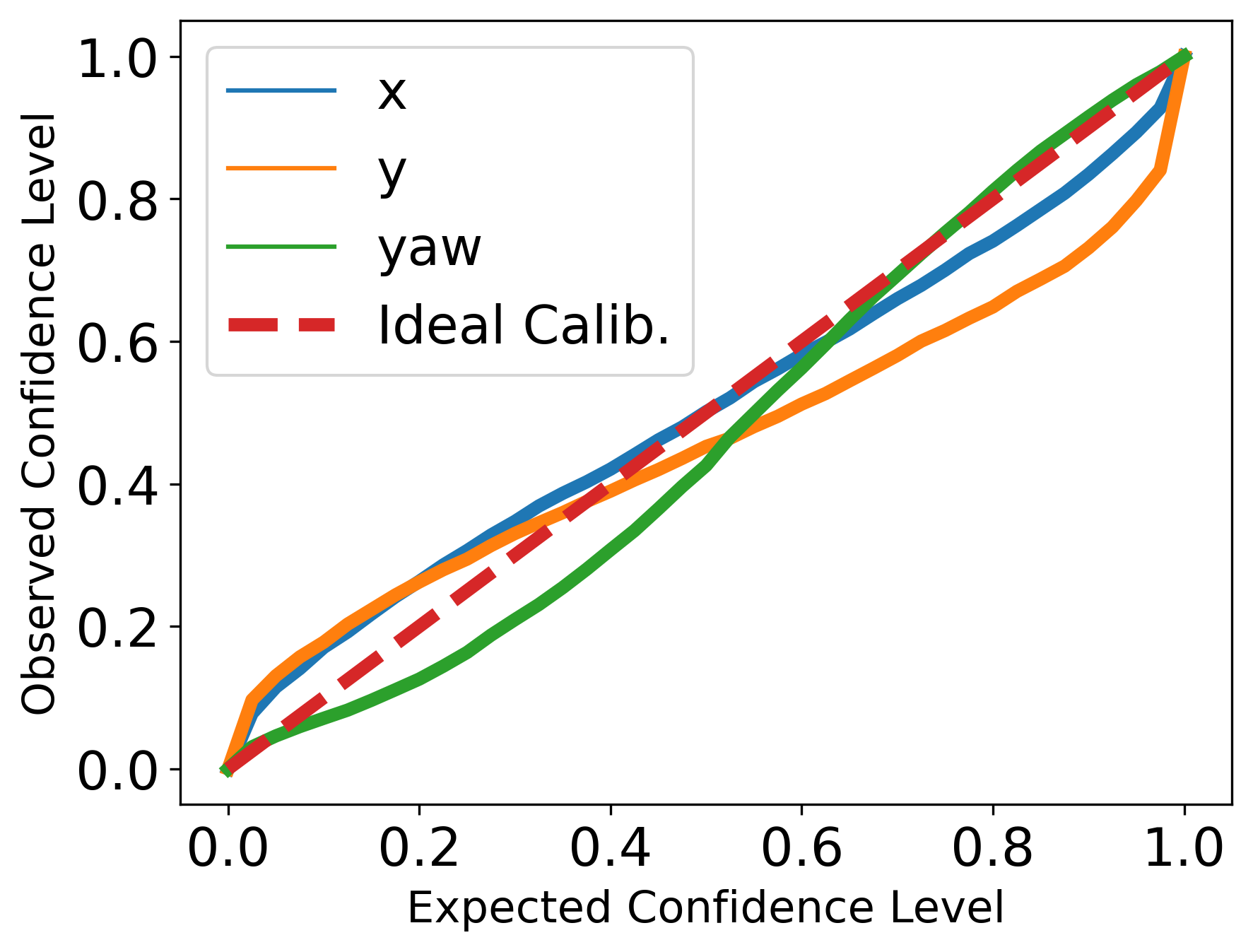}
         \caption{Deep Evidential Regression}
         \label{fig:cal_evd}
     \end{subfigure}
        \caption{\footnotesize{Calibration curves obtained with \ac{MCD} (left), \ac{DE} (center) and \ac{DER} (rights). On the $x$ axis the expected confidence level, on the $y$ axis the observed confidence level. All the approaches show a good calibration with respect to the components considered. For the sake of clarity, we report only the three most important pose parameters, for a ground vehicle, $x$, $y$, and $yaw$.}}
        \label{fig:cal}
        \vspace{-0.6cm}
\end{figure*}

\subsection{Introduction to CMRNet}
CMRNet is a regression \ac{CNN} used to estimate the 6DoF pose of a camera mounted on-board a vehicle navigating within a \ac{LIDAR} map \cite{Cattaneo_2019}.
In particular, this model takes two different images as input: an RGB image and a \ac{LIDAR} image obtained by synthesizing the map as viewed from an initial rough camera pose estimate $H_{init}$. CMRNet performs localization by implicitly matching features extracted from both images, and estimates the misalignment $H_{out}$ between the initial and the camera pose.

In particular, $H_{out}$ is computed as: $tr_{(1, 3)} = (x, y, z)$ for translations, and unit quaternion $q_{(1, 4)} = (q_x, q_y, q_z, q_w)$ for rotations.  We propose to estimate its epistemic uncertainty by providing a reliability value for each pose component. The estimation of possible cross-correlations between the pose components has not been considered in this paper.

\subsection{Uncertainty-Aware CMRNet}
We define an input camera image with $\mathcal{I}_{c}$, an input LiDAR image as $\mathcal{I}_{l}$, a set of trained weights with $\mathcal{W}$ and an Uncertainty Aware (UA) version of CMRNet as a function $f(\mathcal{I}_{c}, \mathcal{I}_{l}, \mathcal{W})$.

\noindent \emph{\textbf{Monte Carlo Dropout}}:
The idea behind \ac{MCD} is to sample from a posterior distribution by providing different output estimates given a single input, which are later used for computing the mean and variance of a Gaussian distribution. 
This sampling is performed by randomly deactivating the weights of the fully-connected layers using a random dropout function $d(\mathcal{W}, p)$ multiple times during model inference, where $p$ represents the dropout probability. Therefore, for \ac{MCD} there is no modification of the network architecture.
We applied the dropout to the regression part of the original CMRNet architecture. When many correlations between RGB and \ac{LIDAR} features are found, we expect to obtain similar samples, despite the dropout application, that is, we expect our model to be more confident with respect to its predictions. For each pose parameter $\mu_i$, we compute the predicted value and the corresponding epistemic uncertainty as follows:
\begin{equation}
\begin{aligned}
    & \e{\mu_i} = \frac{1}{n} \cdot \sum_{n}{f(\mathcal{I}_{c}, \mathcal{I}_{l}, d_{regr}(\mathcal{W}, p))}, 
    \\
    & Var[\mu_i] = \frac{1}{n} \cdot \sum_{n}{(f(\mathcal{I}_{c}, \mathcal{I}_{l}, d_{regr}(\mathcal{W}, p)) - \e{\mu_i})^2}
\end{aligned}
\end{equation}
where $n$ is the number of samples drawn for a given input. Please note that $\e{\mu_i}$ and $Var[\mu_i]$, for the orientation, are computed after the conversion from unit quaternion to Euler angles.

\noindent \emph{\textbf{Deep Ensemble}}: 
\ac{DE}-based approaches perform posterior sampling by exploiting different models trained using different initialization of the weights, but sharing the same architecture.

Using different parameterizations of the same model leads to the recognition of a wider range of data-patterns, and to an increment of the overall accuracy \cite{fort2019deep}. On the other hand, when receiving in input patterns not well-represented in the training set, all the \ac{NN}s in the ensemble would give out low-quality results, so leading to an increment of variance.
In our case, we expect to obtain large epistemic uncertainty when each model identifies a different set of correspondences between RGB and \ac{LIDAR} features, leading to significant different pose estimates.
By training CMRNet $n$ times with different random initializations, we obtain a set of weights $\mathcal{W}_{set} = \{\mathcal{W}_1, ..., \mathcal{W}_n\}$, which describe different local minima of the model function $f(\cdot)$. For each pose parameter $\mu_i$ we compute the predicted expected value and the corresponding epistemic uncertainty as follows:
\begin{equation}
\begin{aligned}
    & \e{\mu_i} = \frac{1}{n} \cdot \sum^{n}_{j=1}{f(\mathcal{I}_{c}, \mathcal{I}_{l}, \mathcal{W}_j)}, 
    \\
    & Var[\mu_i] = \frac{1}{n} \cdot \sum^{n}_{j=1}{(f(\mathcal{I}_{c}, \mathcal{I}_{l}, \mathcal{W}_j) - \e{\mu_i})^2}
\end{aligned}
\end{equation}
where $n$ represents the number of models of the ensemble. In this case too, $\e{\mu_i}$ and $Var[\mu_i]$ of rotations are computed after the conversion from unit quaternion to Euler angles. 

\noindent \emph{\textbf{Deep Evidential Regression}}:
While adapting to \ac{MCD} and \ac{DE} methods does not require particular modifications of CMRNet, the technique proposed by Amini \etal \cite{Amini_2020} requires substantial changes both in the training procedure and in the final part of the architecture.
In Deep Evidential Regression, the main goal is to estimate the parameters of a  Normal Inverse Gamma distribution $NIG(\gamma, \nu, \alpha, \beta)$.
A neural network is trained to estimate the \ac{NIG} parameters, which are then used to compute the expected value and the corresponding epistemic uncertainty, for each pose parameter:
\begin{equation}
    \e{\mu} = \gamma, \qquad
    Var[\mu] = \frac{\beta}{\nu (\alpha - 1)}
\end{equation}
To train the model, the authors propose to exploit the Negative Log Likelihood $\mathcal{L}^{NLL}$ and the Regularization $\mathcal{L}^{R}$ loss functions to maximize and regularize evidence:
\begin{equation}
    \mathcal{L}(\mathcal{W}) = \mathcal{L}^{NLL}(\mathcal{W}) + \lambda \cdot \mathcal{L}^{R}(\mathcal{W})
\end{equation}
\begin{equation}
    \mathcal{L}^{NLL} = - \log p(y | m) \qquad \mathcal{L}^{R} = \Phi \cdot |y - \gamma|
\end{equation}
where $\Phi = 2\nu + \alpha$ is the amount of evidence, see \cite{Amini_2020} for details, and $\lambda$ represents a manually-set parameter that affects the scale of uncertainty, $p(y | m)$ represents the likelihood of the NIG. Note that, $p(y | m)$ is a \emph{pdf} that follows a t-Student distribution $St(\gamma, \frac{\beta(1 +\nu)}{\nu \alpha}, 2\alpha)$ evaluated with respect to a target $y$.
For a complete description of loss functions and theoretical aspects of \ac{DER}, please refer to the work of Amini \etal \cite{Amini_2020}.
To integrate \ac{DER} within CMRNet, we need to deal with the following issues: how to apply \ac{DER} for regressing multiple parameters, how to manage rotations, and how to aggregate the results when computing the final loss.
We changed the last FC-layers, which predict the rotation $q_{(1, 4)} = (q_x, q_y, q_z, q_w)$  and translation $tr_{(1, 3)} = (x, y, z)$ components, in order to estimate the \ac{NIG} distributions associated to each pose parameter.  
As it can be seen in Fig. \ref{fig:cmrnet_der}, we modified CMRNet to regress Euler angles instead of quaternions, then we changed the FC-layers to produce the matrices $eul_{(4, 3)}$ and $tr_{(4, 3)}$, where each column $|\gamma_i, \nu_i, \alpha_i, \beta_i|'$ represents a specific \ac{NIG} \cite{Amini_2020}.
Since the original CMRNet model represents rotations using unit quaternions $q_{(1, 4)}$, we cannot compute the $\mathcal{L}^{NLL}$ and $\mathcal{L}^{R}$ loss functions directly, as addition and multiplication have different behavior on the $S^3$ manifold. As mentioned above, we modified the last FC-layer of CMRNet to directly estimate Euler angles $eul_{(1, 3)} = (r, p, y)$. We also substitute the quaternion distance-based loss used in \cite{Cattaneo_2019} with the smooth $\mathcal{L}_1$ loss \cite{Girshick_2015_ICCV}, which will be later used also in $\mathcal{L}^R$ and $\mathcal{L}^D$, by also considering the discontinuities of Euler angles. Although the Euler angles representation is not optimal \cite{Schneider_2017}, it allows for  easier management of the training procedure and enables a direct comprehension of uncertainty for rotational components. As we will demonstrate in Sec. \ref{sec:experimental_results}, this change does not produce a significant decrease in accuracy.
Since CMRNet performs multiple regressions, it is necessary to establish an aggregation rule for the $\mathcal{L}^{NLL}$ and $\mathcal{L}^R$ loss functions, which are computed for each predicted pose parameter.
With the application of the original loss as in \cite{Amini_2020} we experienced unsatisfactory results. We are under the impression that, in our task, $\mathcal{L}^{NLL}$ presents an undesirable behavior: since the negative logarithm function is calculated over a probability density, it is not lower bound, as the density gets near to be a delta.
\begin{table}[]
\caption{Localization Results}
\label{tab:loc_res}
\centering
\renewcommand*{\arraystretch}{1.1}
\begin{tabular}{l|cc|cc}
\hline
\multicolumn{1}{c|}{\multirow{2}{*}{\textbf{Method}}} & \multicolumn{2}{c|}{\textbf{\begin{tabular}[c]{@{}c@{}}Translation Error (m)\end{tabular}}} & \multicolumn{2}{c}{\textbf{\begin{tabular}[c]{@{}c@{}}Rotation Error (deg)\end{tabular}}} \\ \cline{2-5} 
\multicolumn{1}{c|}{}                                 & median                                      & mean/std                                            & median                                     & mean/std                                           \\ \hline
Rough Initial Pose & 1.88 & 1.82 ± 0.56 & 9.8 & 9.6 ± 2.8 \\
CMRNet (no iter)                                                & 0.52                                        & 0.65 ± 0.45                                       & 1.3                                       & 1.6 ± 1.2                                      \\
CMRNet + MCD                                          & 0.58                                        & 0.69 ± 0.44                                       & 1.8                                       & 2.1 ± 1.3                                      \\
CMRNet + DE                                           & 0.47                               & 0.57 ± 0.39                              & 1.2                              & 1.5 ± 1.1                             \\
CMRNet + DER                                          & 0.54                                        & 0.65 ± 0.46                                       & 1.8                                       & 2.1 ± 1.4                                     
\end{tabular}
\caption*{\footnotesize{Localization results of different CMRNet versions. We present the results of the original model without any iterative refinement (no iter), but the same strategy proposed in \cite{Cattaneo_2019} could be applied to all the other methods. Note that, we do not alter CMRNet accuracy with out \ac{DER}-based approach.}}
\vspace{-2.6\baselineskip}
\end{table}
We propose to overcome the previous issues by avoiding the computation of the logarithm and considering a distance function that is directly based on the probability density $p(y | m)$, that is the pdf of the t-Student distribution.
Therefore, we replaced $\mathcal{L}^{NLL}$ with the following loss $\mathcal{L}^D$ and we also reformulate $\mathcal L^R$:
\begin{equation}
    \mathcal{L}^D =\frac{1}{n} \cdot \sum_{i=1}^{n} d(p(y_i | m_i)^{-1}, 0)
    \quad
    \mathcal{L}^R = \frac{1}{n} \cdot \sum_{i=1}^{n} d(y_i, \gamma_i) \cdot \Phi_i 
\end{equation}
Similarly to $\mathcal{L}^{NLL}$, the idea behind $\mathcal{L^D}$ is to penalize predictions according to the confidence level output by our model with respect to the deviation between a target and an estimated values. However, since this loss function admits a lower bound and is defined in the positive interval, it allows direct computation of a distance metric $d(\cdot)$ on the vector of inverse densities.
To ensure a better numerical stability, we clip $p(y_i | m_i)$ when it returns too low density values, \ie{$<0.04$}.
Regarding $\mathcal{L}^R$, we simply scale the distance error on each pose component with the respective evidence. We the compute the mean error by managing rotations and translations separately. 
The final evidence loss is computed as follows:
\begin{equation}
    \mathcal{L}^{evd} = \mathcal{L}^D + \lambda \mathcal{L}^R
\end{equation}

We noticed that the localization accuracy was decreasing, when employing only $\mathcal{L}^{evd}$ during training. Therefore, we opted to also employ the original geometric loss function $\mathcal{L}^{G}_{tr}$ used in \cite{Cattaneo_2019}, and to employ the smooth $L1$ loss on rotations as geometric loss $\mathcal{L}^{G}_{rot}$.

The overall loss is therefore computed as follows:
\begin{equation}
    \mathcal{L}_{rot} = \mathcal{L}^{G}_{rot} + s^{evd}_{rot} \cdot \mathcal{L}^{evd}_{rot}
\qquad
    \mathcal{L}_{tr} = \mathcal{L}^{G}_{tr} + s^{evd}_{tr} \cdot \mathcal{L}^{evd}_{tr}
\end{equation}
\begin{equation}
    \mathcal{L}_{final} = s_{rot} \cdot \mathcal{L}_{rot} + s_{tr} \cdot \mathcal{L}_{tr}
\end{equation}
where the $s$ hyper-parameters represent scaling factors.

\begin{table}[]
\scriptsize
\caption{Ablation study - CMRNet + DER}
\label{tab:ablation}
\centering
\renewcommand*{\arraystretch}{1.25}
\begin{tabular}{c@{\hspace{0.5\tabcolsep}} c@{\hspace{0.5\tabcolsep}} c|cc|cc}
\hline
\multicolumn{1}{l}{\multirow{2}{*}{$\mathcal{L}^{evd}$}} & \multirow{2}{*}{$\mathcal{L}^G$} & \multirow{2}{*}{$s^{evd}$} & \multicolumn{2}{c|}{\begin{tabular}[c]{@{}c@{}}\textbf{Loc. Error (mean/std)}\end{tabular}} & \multicolumn{2}{c}{\begin{tabular}[c]{@{}c@{}}\textbf{Calib. Error (mean/std)}\end{tabular}} \\ \cline{4-7} 
\multicolumn{1}{l}{}                           &                          &                         & \textbf{Tr. (m)}                                & \textbf{Rot. (°)}                                & \textbf{Tr.}                                     & \textbf{Rot.}                                    \\ \hline
$\mathcal{L}^{NLL}$                                            & -                      & $1.$                     & 1.23 ± 0.57                                    & 2.0 ± 1.7                                     & .080 ± .069                                    & .135 ± .082                                     \\
$\mathcal{L}^{D}$                                            & -                      & $1.$                     & 0.91 ± 0.53                                    & 2.6 ± 1.5                                     & .041 ± .041                                     & .080 ± .074                                     \\
$\mathcal{L}^{NLL}$                                            & \checkmark                      & $1e^{-1}$                    & 0.90 ± 0.56                                    & 1.8 ± 1.4                                     & .090 ± .056                                     & .172 ± .120                                     \\
$\mathcal{L}^{D}$                                            & \checkmark                      & $1e^{-1}$                     & 074 ± 0.49                                   & 2.5 ± 1.4                                     & .035 ± .027                                     & .093 ± .079                                     \\
$\mathcal{L}^{NLL}$                                            & \checkmark                      & $5e^{-3}\dagger$                     & 0.68 ± 0.49                                   & 1.7 ± 1.3                                     & .107 ± .073                                     & .150 ± .010                                     \\
$\mathcal{L}^{D}$                                            & \checkmark                      & $5e^{-3}\dagger$                     & 0.65 ± 0.46                                   & 2.1 ± 1.4                                     & .063 ± .040                                     & .076 ± .060                                    
\end{tabular}
\caption*{\footnotesize{$\dagger$ is the two training steps procedure described in section \ref{sec:method}C.}}
\vspace{-2.5\baselineskip}
\end{table}

\subsection{Training Details}
For all three methods (\ie{\ac{MCD}, \ac{DE}, \ac{DER}}), we followed a similar training procedure as in \cite{Cattaneo_2019}. We trained all models from scratch for a total of $400$ epochs, by fixing a learning rate of $1e^{-4}$, by using the ADAM optimizer and a batch size of 24 on a single NVidia GTX1080ti. The code was implemented with the PyTorch library \cite{pytorch_2019}.
Concerning the \ac{DE} models, random weights initialization was performed by defining a random seed before each training.
For \ac{DER} we initially fixed the scaling parameters $(s_{rot}, s_{tr}, \lambda_{rot} \lambda_{tr}) = (1., 1., 0.01, 0.1)$ and $(s_{rot}^{evd}, s_{tr}^{evd}) = (0.1, 0.1)$. However, we experienced an increment of $\mathcal{L}^{evd}$ after approximately $150$ epochs. Therefore, we decided to stop the training, change $(s_{rot}^{evd}, s_{tr}^{evd}) = (5e^{-3}, 5e^{-3})$, and then proceed with the training. This modification mitigated overfitting. Deactivating $\mathcal{L}^{evd}$ during the second training step led to uncalibrated uncertainties.

\section{EXPERIMENTAL RESULTS}\label{sec:experimental_results}

The experimental activity described in the following section has a dual purpose. On the one hand, it proves that the localization performances of the proposed models achieve comparable results concerning the original CMRNet implementation, providing at the same time reliable uncertainty estimates. On the other hand, we propose one possible application of the estimated uncertainties through a rejection scheme for the vehicle localization problem.

\begin{table}[]
\centering
\caption{Mean Calibration Errors}
\label{tab:calib_err}
\renewcommand*{\arraystretch}{1.1}
\begin{tabular}{lccc}
\hline
\textbf{Axis}            & \textbf{CMRNet+MCD}  & \textbf{CMRNet+DE}   & \textbf{CMRNet+DER}  \\ \hline
\multicolumn{1}{l|}{\textbf{x}}     & 0.045 ± 0.025          & 0.077 ± 0.040        & \textbf{0.042 ± 0.023} \\
\multicolumn{1}{l|}{\textbf{y}}     & \textbf{0.066 ± 0.032}          & 0.093 ± 0.056        & 0.081 ± 0.052         \\
\multicolumn{1}{l|}{z}     & 0.148 ± 0.082          & 0.062 ± 0.036        & 0.067 ± 0.027          \\ \hline
\multicolumn{1}{l|}{roll}  & 0.126 ± 0.069            & 0.068 ± 0.033      & 0.080 ± 0.043          \\
\multicolumn{1}{l|}{pitch} & 0.162 ± 0.092          & 0.050 ± 0.041          & 0.106 ± 0.063 \\
\multicolumn{1}{l|}{\textbf{yaw}}   & 0.069 ± 0.049           & 0.089 ± 0.057          & \textbf{0.042 ± 0.035}         
\end{tabular}
\vspace{-1.6\baselineskip}
\end{table}

\subsection{Dataset}

We used the KITTI odometry dataset \cite{Geiger2012CVPR} to train and validate our models, following the experimental setting proposed in \cite{Cattaneo_2019}. 
In particular, we used images and LiDAR data from KITTI sequences $03$ to $09$, and sequence $00$ for the assessment of the estimated-uncertainty quality. Run 00 presents a negligible overlap of approximately $4\%$ compared to the other sequences, \ie{resulting in a fair validation containing a different environment never seen by CMRNet at training time}.
We exploited the ground truth poses provided by \cite{behley2019iccv} to create accurate \ac{LIDAR} maps.
To simulate the initial rough pose estimate, we added uniformly distributed noise both on translation $[-2m; +2m]$ and rotation components $[-10^{\circ}; +10^{\circ}]$.
To mimic real-life usage and differently from \cite{Cattaneo_2019}, we removed all dynamic objects (\eg{cars and pedestrians}) from within the \ac{LIDAR} maps, allowing some mismatches between the RGB image and the \ac{LIDAR} image. This aspect makes the task more difficult since now CMRNet has also to implicitly learn how to discard incorrect matches.

\subsection{Evaluation metrics}

We evaluated the proposed methods by comparing both localization estimates and uncertainty calibration accuracies. 
In particular, we assessed the localization by measuring the euclidean and quaternion distances between the ground truth and the estimated translation/rotation components. Note that, differently from \cite{Cattaneo_2019}, our main goal is not to minimize the localization error. Instead, we aim to provide a reliability estimate by means of epistemic uncertainty estimation without undermining CMRNet performance.
In particular, we verified the accuracy of the estimated uncertainty using the calibration curves proposed by Kuleshov \etal \cite{pmlr-v80-kuleshov18a}.
This procedure allows us to reveal whether the trained model produces inflated or underestimated uncertainties, by comparing the observed and the ideal confidence level. 

\subsection{Localization assessment}

Our experimental activities encompass the evaluation of the localization performances using all the methods presented in Section 3B, with respect to the original CMRNet proposal. 
Concerning CMRNet + MCD, we applied the dropout to the FC layers with a probability of $0.3$ and obtained the approximated epistemic uncertainty by exploiting $30$ samples. 
Our extensive experimental activity proves this setting provides the best trade-off between accuracy, uncertainty calibration, and computational time.
We implemented a similar approach to identify the suitable number of networks as regards the CMRNet + DE approach. Here we identified the best performances in using $5$ networks, not noticing any performance gain by adding more models to the ensemble.
Table \ref{tab:loc_res} shows the obtained  localization results, together with the statistics of the initial rough pose distribution.
MCD decreases the performances of the original CMRNet, resulting in the worst method among those evaluated. 
On the other hand, CMRNet + DE achieves the best results in terms of accuracy, at the expense of having to train and execute $n$ different networks. This method reduces the errors' standard deviation, as expected from ensemble-based method.
Lastly, CMRNet + DER achieves results comparable to the original CMRNet implementation, proving that our modifications had any negative effect in terms of accuracy.
Table \ref{tab:ablation} reports a brief ablation study performed to find the optimal training parameterization from which we obtained the best DER-based model (last row).

\subsection{Uncertainty Calibration}
The quality of the uncertainty estimates, \ie{the mean calibration errors for the translation and rotation components}, are reported in Table \ref{tab:calib_err}. The errors represent the mean distances between the ideal (\ie{$y = x$}) and the observed calibration, for each confidence interval.
Furthermore, in Fig. \ref{fig:cal}  we show the calibration curves of the most relevant pose parameters.
All three methods obtain good uncertainty calibration, \ie they provide realistic quantities. However, CMRNet + DER shows a better performance in terms of mean calibration errors, considering the most important pose parameters for a ground vehicle (x, y, and yaw). Having a well-calibrated uncertainty-aware model with normal distributions has a major advantage, as its realistic uncertainty estimates can be employed within error filtering algorithms, such as Kalman filters.

\begin{table}[]
\centering
\caption{Localization Results - Discarded Predictions}
\label{tab:loc_top}
\renewcommand*{\arraystretch}{1.1}
\begin{tabular}{l|ccl|cc|c}
\hline
\multicolumn{1}{c|}{\multirow{2}{*}{\textbf{Method}}} & \multicolumn{3}{c|}{\textbf{Translation Error (m)}}    & \multicolumn{2}{c|}{\textbf{Rotation Error (deg)}} & \multirow{2}{*}{\textbf{\begin{tabular}[c]{@{}c@{}}Discarded\\ Pred.\end{tabular}}} \\ \cline{2-6}
\multicolumn{1}{c|}{}                                 & median              & \multicolumn{2}{c|}{mean/std}        & median                 & mean/std                      &                                                                                    \\ \hline
MCD                                                   & 0.58                & \multicolumn{2}{c|}{0.68 ± 0.43} & 1.7                    & 2.0 ± 1.2                 & 27.2\%                                                                               \\
DE                                                    & 0.42                & \multicolumn{2}{c|}{0.50 ± 0.32} & 1.1                    & 1.3 ± 0.8                  & 24.7\%                                                                               \\
DER                                                   & 0.49                & \multicolumn{2}{c|}{0.58 ± 0.38} & 1.6                    & 1.9 ± 1.1                 & 22.0\%                                                                              
\end{tabular}
\vspace{-1.6\baselineskip}
\end{table}

\subsection{Inaccurate Predictions Detection}
\begin{figure}
         \centering
         \includegraphics[width=.475\textwidth]{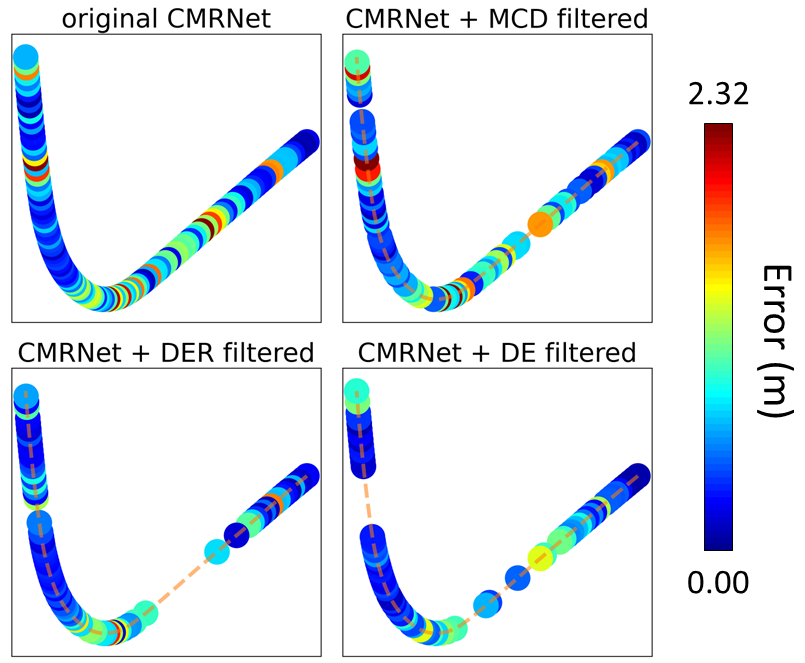}
         \caption{\footnotesize{Qualitative comparison between original CMRNet and our uncertainty aware models on a slice of the kitti 00 run. While the original CMRNet provides inaccurate estimates in the proximity of the depicted curve, CMRNet + DE and CMRNet + DER are able to identify localization failures and finally to discard them.}}
         \label{fig:debugging}
         \vspace{-1.6\baselineskip}
\end{figure}

By measuring the calibration we test the ability of an uncertainty estimator to produce realistic uncertainties. However, we still need to prove a direct proportion between the \ac{DNN} prediction error and the corresponding uncertainty degree.
Besides offering realistic uncertainty estimates, an uncertainty-aware model should assign a large uncertainty to an inaccurate prediction \cite{Amini_2020}.
For instance, a higher level algorithm could exploit a CMRNet estimate according to its associated uncertainty, \eg{by deciding whether to rely only on the measure provided by a \ac{GNSS} or even the subsequent correction performed by the \ac{CNN}}.
To assess that our model provides large uncertainties in presence of very inaccurate predictions, we introduce the following threshold-based strategy. 
For both translation and rotation, we compute the trace of the covariance matrix and compare them to a threshold that allows us to discard predictions with large uncertainty. Rather than deciding an arbitrary value for the thresholds, we use the value at the top 15\% of the traces of the entire validation set, respectively for translation and rotation. The prediction is therefore discarded when both the trace of the covariance of the translation and of the covariance of the rotation are larger than their threshold.
\begin{figure}
         \centering
         \includegraphics[width=.47\textwidth]{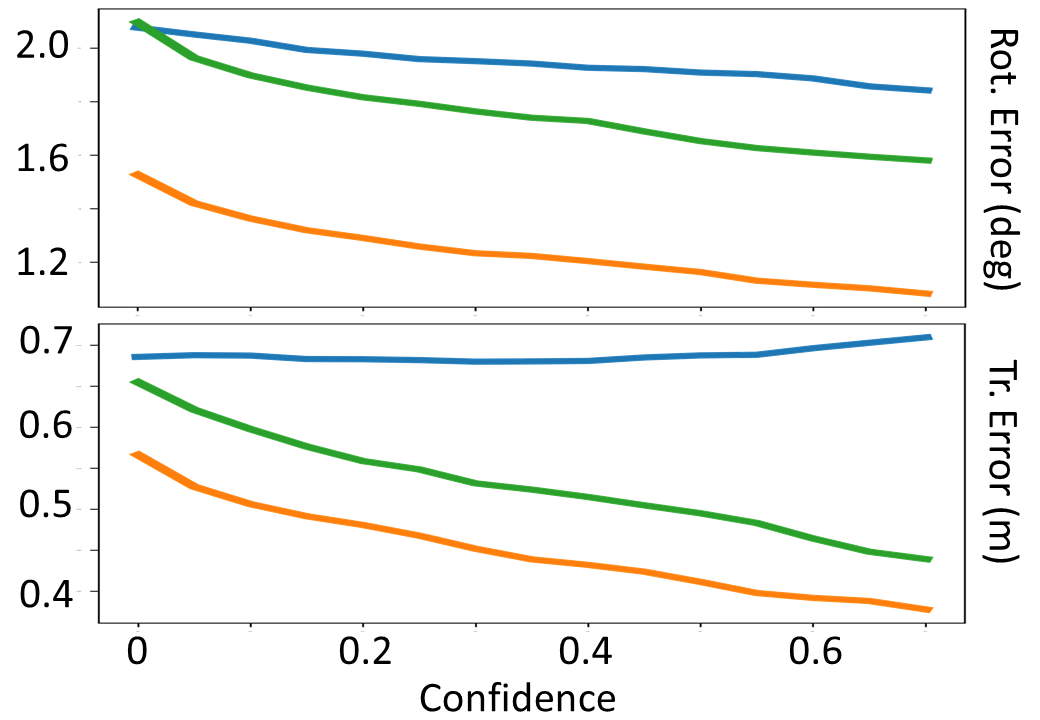}
         \caption{\footnotesize{Prediction errors \emph{vs} CMRNet confidence level. High confidence coincides with small uncertainty (except for MCD). Blue color corresponds to MCD, orange to DE, and green to DER. With DE and DER we can assign large uncertainty to inaccurate predictions.}}
         \label{fig:unc-decr}
         \vspace{-1.55\baselineskip}
\end{figure}
In Table \ref{tab:loc_top} we report the translation and rotation errors, together with the percentage of discarded predictions from a total of $4541$ frames. As can be seen, with CMRNet + DE we are able to detect inaccurate estimates and improve the overall accuracy. With CMRNet + DER we obtain a large localization improvement, outperforming the original model. Furthermore, CMRNet + DER discards fewer  predictions than the other methods, which means that it is able to produce more consistent uncertainties with respect to the different pose components.
Although CMRNet + MCD provides good uncertainty calibration, this model is not able to produce uncertainty estimates that increase with the prediction accuracy. In fact, we obtain the same localization results reported in Table \ref{tab:loc_res} even though such a method discards the largest amount of samples.
In Fig. \ref{fig:unc-decr}, we report the localization accuracy of each proposed method by varying the top$\%$ threshold used for discarding predictions. As can be seen, when the model confidence increases (low uncertainty), its accuracy increases as well.
Another advantage of CMRNet + DE and CMRNet + DER is shown in Fig. \ref{fig:debugging}. Each plot represents the same piece of the path ($125$ frames) of the KITTI 00 run; in this curve, all methods show large localization errors. However, by exploiting \ac{DE} and \ac{DER} we are able to detect most localization failures. This is an interesting property since both \ac{DE} and \ac{DER} can also be exploited as a tool to discover in which scenes CMRNet is likely to fail, even for datasets without an accurate pose ground truth.

\section*{CONCLUSIONS}
We proposed an application of state-of-the-art methods for uncertainty estimation in a multi-modal \ac{DNN} for camera localization. 
In particular, we considered two sampling-based methods, \ie \ac{MCD} and \ac{DE} \cite{Gal_2016, Balaji_2017}, and a direct uncertainty estimation approach named \ac{DER} \cite{Amini_2020}.
To evaluate these methods, we proposed to integrate them within CMRNet \cite{Cattaneo_2019}, which performs map-agnostic camera localization by matching a camera observation with a \ac{LIDAR} map. 
The experiments performed on the KITTI dataset evaluate localization accuracy and uncertainty calibration, also assessing the direct proportion between the increase in accuracy and the decrease in the estimated uncertainty. 
Although CMRNet + MCD showed good localization accuracy and uncertainty calibration, it cannot guarantee that in presence of large uncertainty, we also obtain large errors.
Instead, this behavior was noticed using CMRNet + DE, together with an increment in the overall localization accuracy and a decrease in the variance in the error distribution.
Finally, without undermining its original localization accuracy, we applied a DER-based approach to CMRNet showing the ability to provide well-calibrated uncertainties that can be also employed to detect localization failures using a one-shot estimation scheme.
To the best of our knowledge, this is the first work that integrates a \ac{DER}-based approach in a \ac{DNN} for camera pose regression. 

\bibliographystyle{IEEEtran}
\bibliography{IEEEabrv,bibliography}

\begin{thebibliography}{10}
\providecommand{\url}[1]{#1}
\csname url@samestyle\endcsname
\providecommand{\newblock}{\relax}
\providecommand{\bibinfo}[2]{#2}
\providecommand{\BIBentrySTDinterwordspacing}{\spaceskip=0pt\relax}
\providecommand{\BIBentryALTinterwordstretchfactor}{4}
\providecommand{\BIBentryALTinterwordspacing}{\spaceskip=\fontdimen2\font plus
\BIBentryALTinterwordstretchfactor\fontdimen3\font minus
  \fontdimen4\font\relax}
\providecommand{\BIBforeignlanguage}[2]{{%
\expandafter\ifx\csname l@#1\endcsname\relax
\typeout{** WARNING: IEEEtran.bst: No hyphenation pattern has been}%
\typeout{** loaded for the language `#1'. Using the pattern for}%
\typeout{** the default language instead.}%
\else
\language=\csname l@#1\endcsname
\fi
#2}}
\providecommand{\BIBdecl}{\relax}
\BIBdecl

\bibitem{valada_2018}
N.~Radwan, A.~Valada, and W.~Burgard, ``Vlocnet++: Deep multitask learning for
  semantic visual localization and odometry,'' \emph{IEEE Robotics and
  Automation Letters}, vol.~3, no.~4, pp. 4407--4414, 2018.

\bibitem{Sarlin_2021_CVPR}
P.-E. Sarlin, A.~Unagar, M.~Larsson, H.~Germain, C.~Toft, V.~Larsson,
  M.~Pollefeys, V.~Lepetit, L.~Hammarstrand, F.~Kahl, and T.~Sattler, ``Back to
  the feature: Learning robust camera localization from pixels to pose,'' in
  \emph{Proceedings of the IEEE/CVF Conference on Computer Vision and Pattern
  Recognition (CVPR)}, June 2021, pp. 3247--3257.

\bibitem{McAllister_2017}
R.~McAllister, Y.~Gal, A.~Kendall, M.~Van Der~Wilk, A.~Shah, R.~Cipolla, and
  A.~Weller, ``Concrete problems for autonomous vehicle safety: Advantages of
  bayesian deep learning,'' in \emph{Proceedings of the 26th International
  Joint Conference on Artificial Intelligence}, ser. IJCAI'17.\hskip 1em plus
  0.5em minus 0.4em\relax AAAI Press, 2017, p. 4745–4753.

\bibitem{kendall_2017_unc}
\BIBentryALTinterwordspacing
A.~Kendall and Y.~Gal, ``What uncertainties do we need in bayesian deep
  learning for computer vision?'' in \emph{Advances in Neural Information
  Processing Systems}, I.~Guyon, U.~V. Luxburg, S.~Bengio, H.~Wallach,
  R.~Fergus, S.~Vishwanathan, and R.~Garnett, Eds., vol.~30.\hskip 1em plus
  0.5em minus 0.4em\relax Curran Associates, Inc., 2017. [Online]. Available:
  \url{https://proceedings.neurips.cc/paper/2017/file/2650d6089a6d640c5e85b2b88265dc2b-Paper.pdf}
\BIBentrySTDinterwordspacing

\bibitem{kendall_2016}
A.~Kendall and R.~Cipolla, ``Modelling uncertainty in deep learning for camera
  relocalization,'' in \emph{2016 IEEE International Conference on Robotics and
  Automation (ICRA)}, 2016, pp. 4762--4769.

\bibitem{deng2022deep}
H.~Deng, M.~Bui, N.~Navab, L.~Guibas, S.~Ilic, and T.~Birdal, ``Deep bingham
  networks: Dealing with uncertainty and ambiguity in pose estimation,''
  \emph{International Journal of Computer Vision}, pp. 1--28, 2022.

\bibitem{Cattaneo_2019}
D.~Cattaneo, M.~Vaghi, A.~L. Ballardini, S.~Fontana, D.~G. Sorrenti, and
  W.~Burgard, ``Cmrnet: Camera to lidar-map registration,'' in \emph{2019 IEEE
  Intelligent Transportation Systems Conference (ITSC)}, 2019, pp. 1283--1289.

\bibitem{Amini_2020}
\BIBentryALTinterwordspacing
A.~Amini, W.~Schwarting, A.~Soleimany, and D.~Rus, ``Deep evidential
  regression,'' in \emph{Advances in Neural Information Processing Systems},
  H.~Larochelle, M.~Ranzato, R.~Hadsell, M.~Balcan, and H.~Lin, Eds.,
  vol.~33.\hskip 1em plus 0.5em minus 0.4em\relax Curran Associates, Inc.,
  2020, pp. 14\,927--14\,937. [Online]. Available:
  \url{https://proceedings.neurips.cc/paper/2020/file/aab085461de182608ee9f607f3f7d18f-Paper.pdf}
\BIBentrySTDinterwordspacing

\bibitem{Kendall_2015_ICCV}
A.~Kendall, M.~Grimes, and R.~Cipolla, ``Posenet: A convolutional network for
  real-time 6-dof camera relocalization,'' in \emph{Proceedings of the IEEE
  International Conference on Computer Vision (ICCV)}, December 2015.

\bibitem{Kendall_2017_CVPR}
A.~Kendall and R.~Cipolla, ``Geometric loss functions for camera pose
  regression with deep learning,'' in \emph{Proceedings of the IEEE Conference
  on Computer Vision and Pattern Recognition (CVPR)}, July 2017.

\bibitem{Yin_2018_CVPR}
Z.~Yin and J.~Shi, ``Geonet: Unsupervised learning of dense depth, optical flow
  and camera pose,'' in \emph{Proceedings of the IEEE Conference on Computer
  Vision and Pattern Recognition (CVPR)}, June 2018.

\bibitem{Arandjelovic_2016_CVPR}
R.~Arandjelovic, P.~Gronat, A.~Torii, T.~Pajdla, and J.~Sivic, ``Netvlad: Cnn
  architecture for weakly supervised place recognition,'' in \emph{Proceedings
  of the IEEE Conference on Computer Vision and Pattern Recognition (CVPR)},
  June 2016.

\bibitem{zhu_2018}
\BIBentryALTinterwordspacing
Y.~Zhu, J.~Wang, L.~Xie, and L.~Zheng, ``Attention-based pyramid aggregation
  network for visual place recognition,'' ser. MM '18.\hskip 1em plus 0.5em
  minus 0.4em\relax New York, NY, USA: Association for Computing Machinery,
  2018, p. 99–107. [Online]. Available:
  \url{https://doi.org/10.1145/3240508.3240525}
\BIBentrySTDinterwordspacing

\bibitem{Hausler_2021_CVPR}
S.~Hausler, S.~Garg, M.~Xu, M.~Milford, and T.~Fischer, ``Patch-netvlad:
  Multi-scale fusion of locally-global descriptors for place recognition,'' in
  \emph{Proceedings of the IEEE/CVF Conference on Computer Vision and Pattern
  Recognition (CVPR)}, June 2021, pp. 14\,141--14\,152.

\bibitem{Wolcott_2014}
R.~W. Wolcott and R.~M. Eustice, ``Visual localization within lidar maps for
  automated urban driving,'' in \emph{2014 IEEE/RSJ International Conference on
  Intelligent Robots and Systems}, 2014, pp. 176--183.

\bibitem{Caselitz_2016}
T.~Caselitz, B.~Steder, M.~Ruhnke, and W.~Burgard, ``Monocular camera
  localization in 3d lidar maps,'' in \emph{2016 IEEE/RSJ International
  Conference on Intelligent Robots and Systems (IROS)}, 2016, pp. 1926--1931.

\bibitem{Neubert_2017}
P.~Neubert, S.~Schubert, and P.~Protzel, ``Sampling-based methods for visual
  navigation in 3d maps by synthesizing depth images,'' in \emph{2017 IEEE/RSJ
  International Conference on Intelligent Robots and Systems (IROS)}, 2017, pp.
  2492--2498.

\bibitem{Feng_2019}
M.~Feng, S.~Hu, M.~H. Ang, and G.~H. Lee, ``2d3d-matchnet: Learning to match
  keypoints across 2d image and 3d point cloud,'' in \emph{2019 International
  Conference on Robotics and Automation (ICRA)}, 2019, pp. 4790--4796.

\bibitem{Cattaneo_2020}
D.~Cattaneo, M.~Vaghi, S.~Fontana, A.~L. Ballardini, and D.~G. Sorrenti,
  ``Global visual localization in lidar-maps through shared 2d-3d embedding
  space,'' in \emph{2020 IEEE International Conference on Robotics and
  Automation (ICRA)}, 2020, pp. 4365--4371.

\bibitem{kingma_2015}
\BIBentryALTinterwordspacing
D.~P. Kingma, T.~Salimans, and M.~Welling, ``Variational dropout and the local
  reparameterization trick,'' in \emph{Advances in Neural Information
  Processing Systems}, C.~Cortes, N.~Lawrence, D.~Lee, M.~Sugiyama, and
  R.~Garnett, Eds., vol.~28.\hskip 1em plus 0.5em minus 0.4em\relax Curran
  Associates, Inc., 2015. [Online]. Available:
  \url{https://proceedings.neurips.cc/paper/2015/file/bc7316929fe1545bf0b98d114ee3ecb8-Paper.pdf}
\BIBentrySTDinterwordspacing

\bibitem{Balaji_2017}
\BIBentryALTinterwordspacing
B.~Lakshminarayanan, A.~Pritzel, and C.~Blundell, ``Simple and scalable
  predictive uncertainty estimation using deep ensembles,'' in \emph{Advances
  in Neural Information Processing Systems}, I.~Guyon, U.~V. Luxburg,
  S.~Bengio, H.~Wallach, R.~Fergus, S.~Vishwanathan, and R.~Garnett, Eds.,
  vol.~30.\hskip 1em plus 0.5em minus 0.4em\relax Curran Associates, Inc.,
  2017. [Online]. Available:
  \url{https://proceedings.neurips.cc/paper/2017/file/9ef2ed4b7fd2c810847ffa5fa85bce38-Paper.pdf}
\BIBentrySTDinterwordspacing

\bibitem{Sensoy_2018}
\BIBentryALTinterwordspacing
M.~Sensoy, L.~Kaplan, and M.~Kandemir, ``Evidential deep learning to quantify
  classification uncertainty,'' in \emph{Advances in Neural Information
  Processing Systems}, S.~Bengio, H.~Wallach, H.~Larochelle, K.~Grauman,
  N.~Cesa-Bianchi, and R.~Garnett, Eds., vol.~31.\hskip 1em plus 0.5em minus
  0.4em\relax Curran Associates, Inc., 2018. [Online]. Available:
  \url{https://proceedings.neurips.cc/paper/2018/file/a981f2b708044d6fb4a71a1463242520-Paper.pdf}
\BIBentrySTDinterwordspacing

\bibitem{Meinert_2021}
\BIBentryALTinterwordspacing
N.~Meinert and A.~Lavin, ``Multivariate deep evidential regression,''
  \emph{CoRR}, vol. abs/2104.06135, 2021. [Online]. Available:
  \url{https://arxiv.org/abs/2104.06135}
\BIBentrySTDinterwordspacing

\bibitem{Kendall_2018_CVPR}
A.~Kendall, Y.~Gal, and R.~Cipolla, ``Multi-task learning using uncertainty to
  weigh losses for scene geometry and semantics,'' in \emph{Proceedings of the
  IEEE Conference on Computer Vision and Pattern Recognition (CVPR)}, June
  2018.

\bibitem{Gal_2016}
\BIBentryALTinterwordspacing
Y.~Gal and Z.~Ghahramani, ``Dropout as a bayesian approximation: Representing
  model uncertainty in deep learning,'' in \emph{Proceedings of The 33rd
  International Conference on Machine Learning}, ser. Proceedings of Machine
  Learning Research, M.~F. Balcan and K.~Q. Weinberger, Eds., vol.~48.\hskip
  1em plus 0.5em minus 0.4em\relax New York, New York, USA: PMLR, 20--22 Jun
  2016, pp. 1050--1059. [Online]. Available:
  \url{https://proceedings.mlr.press/v48/gal16.html}
\BIBentrySTDinterwordspacing

\bibitem{Petek_2022}
K.~Petek, K.~Sirohi, D.~Büscher, and W.~Burgard, ``Robust monocular
  localization in sparse hd maps leveraging multi-task uncertainty
  estimation,'' in \emph{2022 International Conference on Robotics and
  Automation (ICRA)}, 2022, pp. 4163--4169.

\bibitem{peretoukhin_2019}
\BIBentryALTinterwordspacing
V.~Peretroukhin, B.~Wagstaff, M.~Giamou, and J.~Kelly, ``Probabilistic
  regression of rotations using quaternion averaging and a deep multi-headed
  network,'' \emph{CoRR}, vol. abs/1904.03182, 2019. [Online]. Available:
  \url{http://arxiv.org/abs/1904.03182}
\BIBentrySTDinterwordspacing

\bibitem{fort2019deep}
S.~Fort, H.~Hu, and B.~Lakshminarayanan, ``Deep ensembles: A loss landscape
  perspective,'' \emph{arXiv preprint arXiv:1912.02757}, 2019.

\bibitem{Girshick_2015_ICCV}
R.~Girshick, ``Fast r-cnn,'' in \emph{Proceedings of the IEEE International
  Conference on Computer Vision (ICCV)}, December 2015.

\bibitem{Schneider_2017}
N.~Schneider, F.~Piewak, C.~Stiller, and U.~Franke, ``Regnet: Multimodal sensor
  registration using deep neural networks,'' in \emph{2017 IEEE Intelligent
  Vehicles Symposium (IV)}, 2017, pp. 1803--1810.

\bibitem{pytorch_2019}
\BIBentryALTinterwordspacing
A.~Paszke, S.~Gross, F.~Massa, A.~Lerer, J.~Bradbury, G.~Chanan, T.~Killeen,
  Z.~Lin, N.~Gimelshein, L.~Antiga, A.~Desmaison, A.~Kopf, E.~Yang, Z.~DeVito,
  M.~Raison, A.~Tejani, S.~Chilamkurthy, B.~Steiner, L.~Fang, J.~Bai, and
  S.~Chintala, ``Pytorch: An imperative style, high-performance deep learning
  library,'' in \emph{Advances in Neural Information Processing Systems 32},
  H.~Wallach, H.~Larochelle, A.~Beygelzimer, F.~d\textquotesingle
  Alch\'{e}-Buc, E.~Fox, and R.~Garnett, Eds.\hskip 1em plus 0.5em minus
  0.4em\relax Curran Associates, Inc., 2019, pp. 8024--8035. [Online].
  Available:
  \url{http://papers.neurips.cc/paper/9015-pytorch-an-imperative-style-high-performance-deep-learning-library.pdf}
\BIBentrySTDinterwordspacing

\bibitem{Geiger2012CVPR}
A.~Geiger, P.~Lenz, and R.~Urtasun, ``Are we ready for autonomous driving? the
  kitti vision benchmark suite,'' in \emph{Conference on Computer Vision and
  Pattern Recognition (CVPR)}, 2012.

\bibitem{behley2019iccv}
J.~Behley, M.~Garbade, A.~Milioto, J.~Quenzel, S.~Behnke, C.~Stachniss, and
  J.~Gall, ``{SemanticKITTI: A Dataset for Semantic Scene Understanding of
  LiDAR Sequences},'' in \emph{Proc. of the IEEE/CVF International Conf.~on
  Computer Vision (ICCV)}, 2019.

\bibitem{pmlr-v80-kuleshov18a}
\BIBentryALTinterwordspacing
V.~Kuleshov, N.~Fenner, and S.~Ermon, ``Accurate uncertainties for deep
  learning using calibrated regression,'' in \emph{Proceedings of the 35th
  International Conference on Machine Learning}, ser. Proceedings of Machine
  Learning Research, J.~Dy and A.~Krause, Eds., vol.~80.\hskip 1em plus 0.5em
  minus 0.4em\relax PMLR, 10--15 Jul 2018, pp. 2796--2804. [Online]. Available:
  \url{https://proceedings.mlr.press/v80/kuleshov18a.html}
\BIBentrySTDinterwordspacing

\end{thebibliography}

\end{document}